# Reply to: Limitations in odour recognition and generalisation in a neuromorphic olfactory circuit

Roy Moyal[1], Nabil Imam[2], Thomas A. Cleland[1*]

[1] Computational Physiology Lab, Department of Psychology, Cornell University, Ithaca, NY, USA
[2] College of Computing, Georgia Institute of Technology, Atlanta, GA USA

* To whom correspondence should be addressed

Dennler et al. [17] submit that they have discovered limitations affecting some of the conclusions drawn in our 2020 paper, *Rapid online learning and robust recall in a neuromorphic olfactory circuit* [1]. Specifically, they assert (1) that the public dataset we used suffers from sensor drift and a nonrandomized measurement protocol, (2) that our neuromorphic EPL network is limited in its ability to generalize over repeated presentations of an odorant, and (3) that our EPL network results can be performance matched by using a more computationally efficient distance measure. Though they are correct in their description of the limitations of that public dataset[2], they do not acknowledge in their first two assertions how our utilization of those data sidestepped these limitations. Their third claim arises from flaws in the method used to generate their distance measure. We respond below to each of these three claims in turn.

## *Metal oxide sensor drift*

We utilized samples of odor responses drawn from an array of 72 metal oxide (MOx) chemosensors spatially dispersed across a wind tunnel [3]. The intended sources of variance in this public dataset include plume dynamics and wide, largely unpredictable variations in odorant concentration at specific sensor sites. Moreover, as noted by Dennler et al., there also are unintended sources of variance embedded in that dataset. To wit: the rate of drift in the response profiles of these MOx sensors generated cumulative changes in sensor responses over the course of data acquisition that are comparable to or greater than the odorant-specific differences in sensor responses on which odorant recognition depends. The fact that these data were acquired in sequential, odorant-specific batches over the course of eight months ensured that these drift-based changes in sensor response profiles would be conflated with the odorant-based differences in sensor responses. Dennler et al. illustrate this with an experiment (their Figs. 1b, S1b) in which they identify odorants based solely on the sensor drift that had accumulated by the day on which a particular odorant was tested.

That said, this problem is not relevant to our findings. We explicitly set aside the problem of MOx sensor drift by randomly selecting single samples from each of the ten odorant responses and occluding them with our own noise models designed to mimic interference by environmental background odorants [1]. While the rate of baseline drift and decay is an existential threat to the utility of MOx-based chemosensor devices, the problem ultimately is associated with these particular materials [4] and can be foreseeably resolved by using different chemosensor technologies [5–9]. Accordingly, we chose to sidestep this MOx-associated problem and instead study the broader problem of identifying odor sources of interest in the presence of unpredictable competitive interference, predicated on reasonably stable sensor responses.

*Generalization*

Dennler *et al.* acknowledge that the EPL network convincingly restores input patterns corrupted by impulse noise (Figs. 3-5 in Ref. 1). Their broad conclusion that the model does not generalize is incorrect; signal restoration *is* a form of generalization and comprises the central message of the paper. Generalization to different samples within bounds also is shown in Fig. 5 of Ref. 1. What they do correctly point out is that a learned odorant representation does not generalize to a separate presentation of the same odorant delivered into the wind tunnel. This requires explanation.

The same odorant, sampled after separate deliveries to these 72 spatially dispersed sensors, would be encountered by each individual sensor at substantially different concentrations according to the random dispersion patterns of each plume. It is well established that concentration effects are more powerful than odorant quality-based differences in their ability to drive MOx sensor responses. Accordingly, their test equates to the addition of high levels of random concentration-based noise to every sensor – high enough to disrupt all odorant-diagnostic pattern information. It is no surprise that Dennler *et al.* failed to identify the test odorant under these conditions.

While it is desirable for a sensor device deployed in the wild to generalize across diverse, statistically ill-behaved odorant presentations, this is not what the EPL network is designed for. This regularization problem is addressed in part by our glomerular layer preprocessor computations [10–13], which specifically incorporate circuitry to provide concentration tolerance and other stabilizing effects. These transformations were excluded from Ref. 1 in order to focus on the capabilities of EPL transformations. Clearly, also, given the goal of odorant recognition, one certainly would not disperse the elements of a multisensor array across a wide area, which randomizes per-sensor analyte concentrations and hence greatly increases the amount of uncorrelated variance without returning any benefit.

*Distance measure performance*

Dennler *et al*. submit that our findings can be "effectively addressed by using a simple hash table", by which they mean a measure of overlap using Jaccard similarity. This claim rests on two errors of analysis. First, the Jaccard similarity coefficients that they report in their Fig. 2 are misleading; their method, based on their own code, provides much lower certainty than that figure indicates (discussed below). Second, because Dennler *et al*. do not employ a similarity threshold θ to determine successful classification, they exclude the possibility of *none of the above* (we used θ = 0.75 in our paper, as noted therein). This forced-choice strategy leads inexorably to false positive results which they interpret as successful classifications. We explain these errors in more detail below.

Dennler *et al.*'s method (code in their *Algorithm 1*) is comparable to the nearest-neighbor algorithm that we use in our paper (Fig. 6a in Ref. 1; *Raw*), except that they employ an inappropriate measure of similarity and do not include a threshold for classification. Briefly, when a 72-element test odorant is presented for classification, it is compared against each of the (ten) 72-element odorant templates. Dennler *et al.*'s algorithm then selects the template with the greatest number of exactly matching values to the test odorant and calculates the Jaccard coefficients between *that selected template* and each of the odorant templates in turn. The outcome of this strategy is that the best match is always reported as a perfect Jaccard similarity of 1.0 (i.e., that of the selected template compared with itself; the actual similarity between the test odorant signature and that of any template is never reported). Each of the five bars corresponding to different odorant templates in the panels of their Fig. 2, then, depicts the median of ten results comprising some number of samples classified as that odorant (each contributing a Jaccard coefficient of 1.0) and some number classified as a different odorant (each contributing the Jaccard coefficient computed between the template of the odorant corresponding to the bar in question and the template of the odorant that the sample was classified as). Among other deficiencies, this practice does not reflect actual classification performance; to wit, if across ten samples of toluene, six are identified as toluene and four as some other odor, a median similarity of 1.0 to toluene is reported, even though four of the ten trials return inaccurate classifications. This strategy is unrelated to the method of our EPL attractor network, in which a test odorant signature is iteratively drawn towards the template(s) of its class, and its progressively increasing similarity to that template is directly reported.

A revealing consequence of Dennler *et al.*'s forced-choice strategy is that even randomly generated vectors, unrelated to any of the learned templates, are classified positively as knowns if they have even a single element that matches that of any template. To test this, we passed 100 randomly generated 72-element vectors to Dennler *et al.*'s algorithm; all were misclassified as knowns with perfect Jaccard similarities of 1.0 to the template of their class (Fig. 1a). An effective remedy for this error, as we presented in Ref. 1, is to define a classification threshold θ such that test odorants are identified as class members only if their similarity to the template exceeds that threshold, otherwise being classified as *none of the above*. Accordingly, the nearest neighbor algorithm that we employed in Ref. 1 (Fig. 6a, *Raw*) classified all 100 random vectors as *none of the above*, whereas our EPL model classified one as known and 99 as *none of the above*.

A converse problem with Dennler *et al.*'s algorithm is that a test odorant may be very closely related to a learned template, but have no precisely identical elements, in which case it is scored as having zero similarity to that learned template. Small amounts of noise can therefore lead Dennler *et al.*'s method to yield effectively random outcomes. To illustrate this, we generated ten test odorants by adding a minimal quantity of noise to each odorant template (+1 to each element). Dennler *et al.*'s algorithm did not correctly classify any of these samples, whereas the Manhattan distance method that we used for benchmarking (Fig. 6 in Ref. 1) accurately classified them all (Fig. 1b). We then modified Dennler *et al.*'s code to use a Manhattan distance-based similarity metric with a detection threshold of θ = 0.75, and found that it then exhibits the same performance (8.2%) as the nearest neighbor algorithm that we utilized in Ref. 1 (Fig. 1c; compare to *Raw* results from Fig. 6a in Ref. 1). The performance benchmarks claimed by Dennler *et al.*, then, arise from an artifactual effect of a similarity metric handcrafted for the specific test samples presented, potentiated by the lack of a threshold requirement and the reporting of all classification decisions as having similarity coefficients of unity.

Finally, Dennler *et al.* report a comparison between the runtime of their code on a CPU and that of our EPL network on the Intel Loihi neuromorphic platform. We do not address that specific comparison here, as their code does not actually solve the problem presented. However, a *k*-nearest-neighbors search algorithm using spike encoding, similar to the problem we solved, was evaluated on a multichip Loihi system; it outperformed a brute-force CPU algorithm by a factor of 685 in energy-delay product [14,15]. Other problem instances using neural networks deployed on Loihi hardware also have exhibited orders of magnitude improvements in runtime and energy efficiency compared to CPU computation [15].

## *Conclusion*

Chemical sensing using MOx sensors is a challenging goal; it is considered an accomplishment for MOx sensor arrays to recognize a natural signal drawn from the same odorous location on the same or next day *without* the addition of any disruptive noise [16]. Rather than wrestle with the limitations of MOx sensors, we seek to improve the back-end intelligence used for the identification of chemical odorants in the presence of competitive/occluding noise and other such challenges, in anticipation of real-world deployment using less problematic chemosensor technologies.

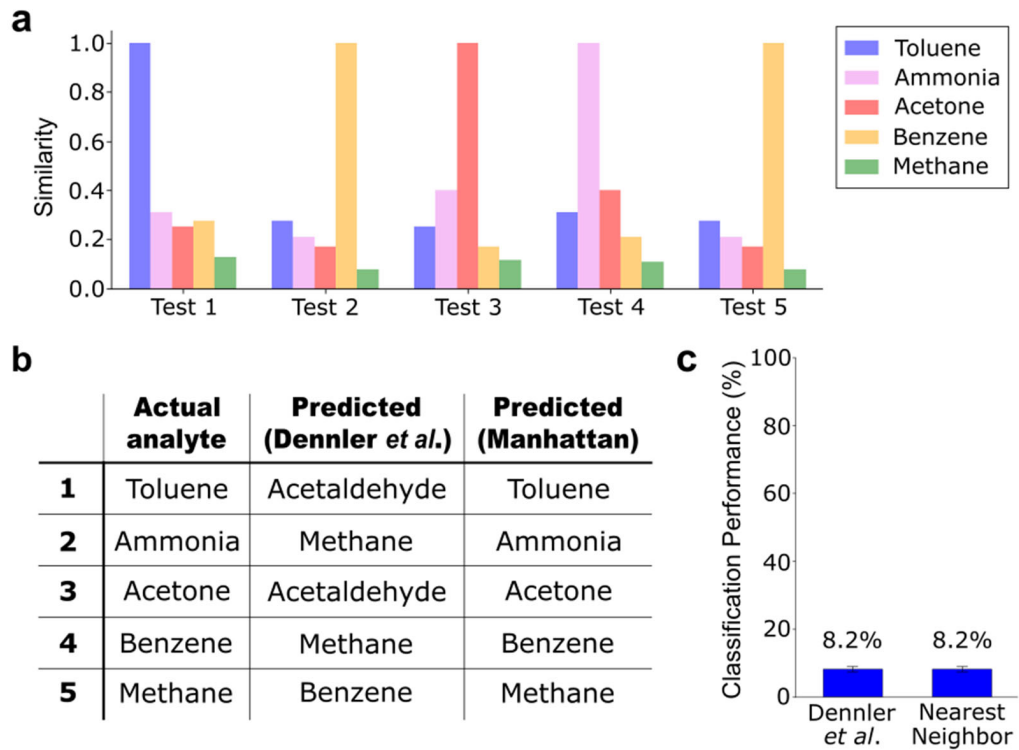

**Fig. 1 | Benchmarking performance of the Dennler *et al*. method. a,** When randomly generated 72-element vectors were presented to Dennler *et al.*'s algorithm, it classified each random vector as a known odorant with Jaccard similarity = 1.0. One hundred independently generated random vectors were presented, all of which were classified as knowns (five of these are depicted). The nearest neighbor algorithm from Ref. 1 classified each of the input vectors as *none of the above*, whereas the EPL algorithm correctly classified 99% as *none of the above*. **b,** We generated ten test samples by adding a small amount of noise (+1) to each of the elements of the ten training odorants. Dennler *et al.*'s algorithm failed to identify any of the test samples correctly (five are depicted). When we modified the algorithm to employ a Manhattan distance-based similarity metric, all of the test samples were classified accurately. **c,** We then further modified Dennler *et al.*'s algorithm to also include a classification threshold of θ = 0.75 to exclude false positives. When we ran the same benchmark utilized in Ref. 1 (100 test samples of each odorant, totaling 1000 samples per simulation, impulse noise occlusion levels randomly selected from the range [0.2-0.8], here reporting the mean of 100 simulations performed with different random seeds), we obtained results identical to the nearest neighbor algorithm from Ref. 1 (Fig. 6a, *Raw*), confirming their equivalence. Error bars depict standard deviations.

### *Data availability*

The gas sensor data referred to herein (Ref. 12) are available from http://archive.ics.uci.edu/dataset/251/gas+sensor+arrays+in+open+sampling+settings

### *Code availability*

Code generating the panels of Fig. 1 is available from doi: 10.6084/m9.figshare.25263661.

### *Author contributions*

All authors contributed to the critical analysis of Dennler *et al.*'s comment on our work, and to the text of the manuscript. RM conducted simulations. TAC compiled and revised the manuscript.

### *Competing interests*

TAC and NI are listed as inventors on US patent US20220198245A1 and European patent EP3953868A4, both pending, on neuromorphic methods for rapid online learning and signal restoration.

### *Support*

This work was supported by NSF grants CBET-2123862 to TAC and EFMA-2223811 to TAC and NI. RM was supported by the Eric and Wendy Schmidt AI in Science Postdoctoral Fellowship, a Schmidt Futures Program.